\documentclass[letterpaper, 10 pt, conference]{ieeeconf} 

\IEEEoverridecommandlockouts                              %

\usepackage{setspace}
\usepackage[margin=2cm]{geometry}  %

\usepackage{multirow} %

\usepackage{microtype}

\usepackage{graphics} %
\usepackage{times} %

\usepackage[font=footnotesize,labelfont=bf,tableposition=top]{caption}
\usepackage{subcaption}
\usepackage{pdfpages}
\usepackage{amsmath,amssymb,amsfonts,amscd}
\usepackage{mathtools}

\usepackage{siunitx} %

\usepackage{siunitx}%

\usepackage{paralist}%

\usepackage[printonlyused,nolist]{acronym} %

\usepackage{algorithm}
\usepackage{algorithmic}

\usepackage{cleveref}

\usepackage{booktabs} %

\makeatletter

\let\NAT@parse\relax
\makeatother

\usepackage[numbers,sort&compress]{natbib} %

\newcommand{\mycomment}[1]{} %

\DeclareMathOperator*{\diag}{diag}

\newcommand{\squeezeWords}{\looseness=-1}

\graphicspath{ {images/} }

\begin{acronym}
    \acro{MVO}[MVO]{multimotion visual odometry}
	\acro{MEP}[MEP]{multimotion estimation problem}
	\acro{MOT}[MOT]{multiobject tracking}
	\acro{SLAM}[SLAM]{simultaneous localizaton and mapping}
	\acro{MVO}[MVO]{multimotion visual odometry}
	\acro{VO}[VO]{visual odometry}
	\acro{AER}[AER]{address-event representation}
	\acro{SAE}[SAE]{surface of active events}
	\acro{KLT}[KLT]{Kanade-Lucas-Tomasi}
	\acro{EKLT}[EKLT]{event-based Kanade-Lucas-Tomas tracker}
	\acro{IMU}[IMU]{inertial measurement unit}
	\acro{VIO}[VIO]{visual inertial odometry}
	\acro{SVO}[SVO]{stereo visual odometry}
	\acro{RPG}[RPG]{robotics and perception group}
	\acro{OMD}[OMD]{Oxford multimotion dataset}
\end{acronym}

\title{\LARGE \bf
Event-based Stereo Visual Odometry  with Native Temporal Resolution \\ via Continuous-time Gaussian Process Regression
}

\author{Jianeng Wang$^{1}$ and Jonathan D. Gammell$^{1}$%
\thanks{$^{1}$Estimation, Search, and Planning (ESP) Group, Oxford Robotics Institute (ORI),
University of Oxford, UK.
{\tt\small\{jianeng, gammell\}@robots.ox.ac.uk.}}
}

\begin{document}

\maketitle
\thispagestyle{empty}
\pagestyle{empty}

\begin{abstract}
Event-based cameras asynchronously capture individual visual changes in a scene.
This makes them more robust than traditional frame-based cameras to highly dynamic motions and poor illumination.
It also means that every measurement in a scene can occur at a unique time. 

Handling these different measurement times is a major challenge of using event-based cameras.
It is often addressed in visual odometry (VO) pipelines by approximating temporally close measurements as occurring at one common time.
This grouping simplifies the estimation problem but, absent additional sensors, sacrifices the inherent temporal resolution of event-based cameras.

This paper instead presents a complete stereo VO pipeline that estimates directly with individual event-measurement times without requiring any grouping or approximation in the estimation state. %
It uses continuous-time trajectory estimation to maintain the temporal fidelity and asynchronous nature of event-based cameras through Gaussian process regression with a physically motivated prior.
Its performance is evaluated on the MVSEC dataset, where it achieves $7.9\cdot 10^{-3}$ and $5.9\cdot 10^{-3}$ RMS relative error on two independent sequences, outperforming the existing publicly available event-based stereo VO pipeline by two and four times, respectively.
\newline
\end{abstract}


\section{INTRODUCTION}
\label{sec:introduction}
Visual Odometry (VO) is a technique to estimate egomotion in robotics \cite{Moravec1980,Nister2004, frame_based_SVO, DSO, SVO}. VO systems using traditional frame-based cameras often struggle in scenarios with high speed motion and poor illumination. In these scenarios, the motion blur and poor image contrast of frame-based cameras result in bad estimation performance.

Event-based cameras perform better than traditional cameras in these challenging scenarios. They detect pixelwise intensity change and report the time at which the change occurs asynchronously. This gives them high temporal resolution and high dynamic range avoiding the limitations of frame-based cameras and providing the potential for more accurate VO systems \cite{e_camera_survey}.

\begin{figure}
  \centering
  \includegraphics[width=1\linewidth]{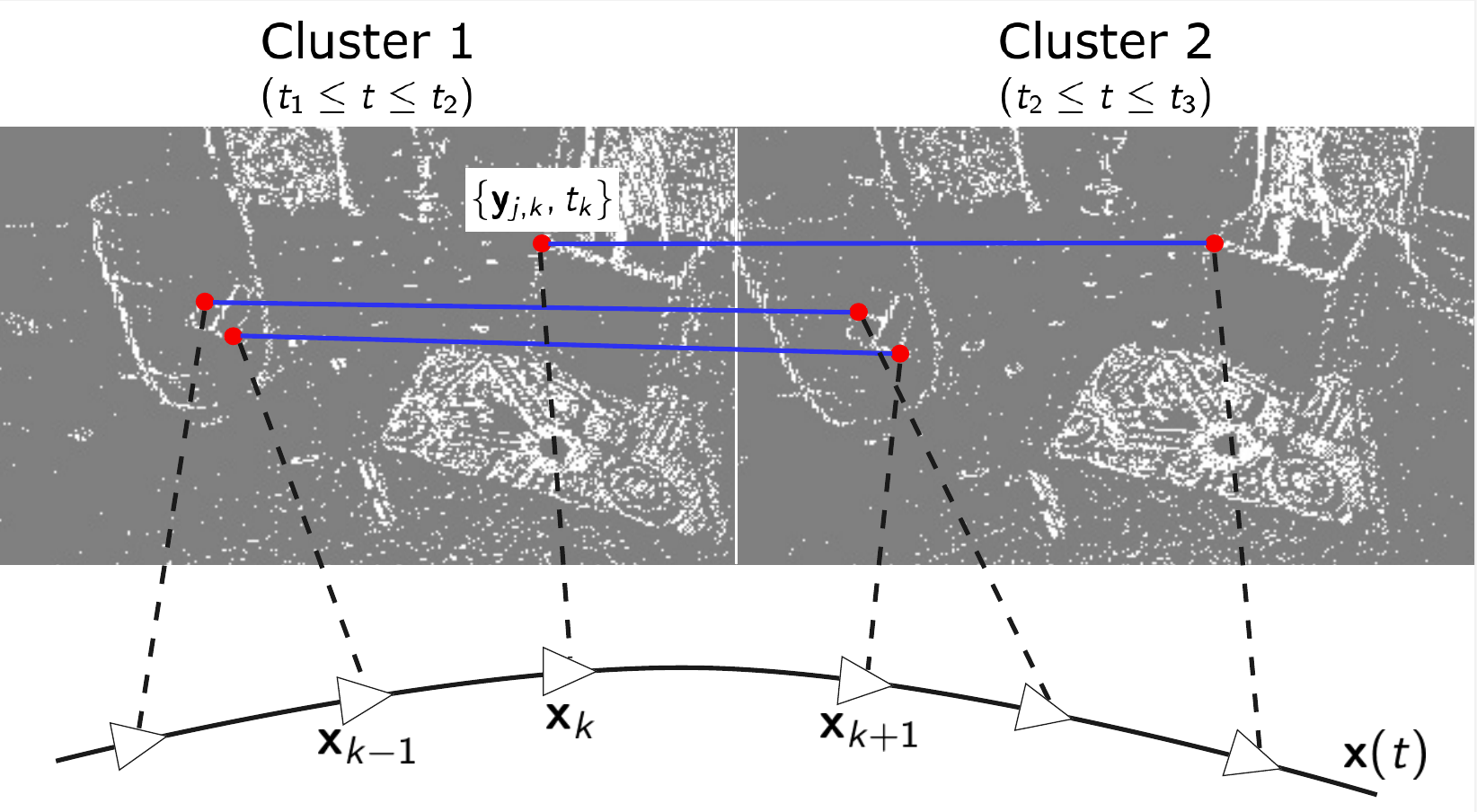}
  \caption{\footnotesize{An illustration of the continuous-time trajectory estimation pipeline. Event clusters are defined by an asynchronous event data stream in discrete windows based on number of events and their times (e.g., $t_1 \leq t \leq t_2$). Features (red) are detected from the resulting clusters and matched with features occurring in other clusters (blue) to create tracklets. Each event feature in the tracklet, $\mathbf{y}_{j,k}$, is a measurement of landmark, $\mathbf{p}_j$, and defines a trajectory state  in the estimation problem, $\mathbf{x}_k = \{\mathbf{T}_{k,1}, \boldsymbol{\varpi}_k\}$, at the measurement time, $t_k$. The camera motion is estimated as a continuous-time trajectory function, $\mathbf{x}(t)$, defined
by the discrete states and a physically founded motion prior.}}
  \label{fig:state_construct}
\end{figure}

Any event-based VO system must address the asynchronous event times. Many pipelines do this by grouping similar feature times to a common time \cite{Ultimate_SLAM, EVO, ESVO}. This allows for the direct application of frame-based VO pipelines but sacrifices the temporal resolution of event cameras.

This paper instead presents an event-based VO system that uses the unique asynchronous timestamps directly in the estimation problem without grouping or approximation. It estimates the camera motion as a continuous-time trajectory represented by states at unique feature times and a white-noise-on-acceleration (WNOA) motion prior. The trajectory is estimated using nonparametric Gaussian process regression. This results in a continuous, physically founded trajectory that exploits the temporal resolution of asynchronous event cameras and can estimate complex, real-world motions. (Fig.~\ref{fig:state_construct}). 
\squeezeWords

This paper presents a complete event-based stereo VO pipeline using continuous-time Gaussian process regression. It is compatible with any feature detector and tracker, including frame-based methods for traditional cameras, without reducing the temporal resolution of event-based cameras. It also uses Motion-Compensated RANSAC (MC-RANSAC) \cite{RANSAC_MC} to consider the unique measurement times during outlier rejection and independently provide better tracklets and initial conditions for estimation. The resulting continuous-time trajectory provides estimates of camera poses at any and all timestamps in the estimation window. 

It is evaluated on the publicly available Multi Vehicle Stereo Event Camera (MVSEC) dataset \cite{MVSEC}, where it obtains a more accurate and smoother trajectory estimate than the state-of-the-art Event-based Stereo Visual Odometry (ESVO) \cite{ESVO}. It achieves $7.9\cdot 10^{-3}$ and $5.9\cdot 10^{-3}$ root-mean-squared (RMS) relative error in $SE(3)$ and 5.78\% and 4.93\%  final global translational error as a percent of path length on two independent MVSEC sequences.
This outperforms the publicly available ESVO on these sequences, especially in terms of trajectory smoothness and RMS relative error where it is two- and four-times better, respectively.

The rest of the paper is organized as follows. Sec. \ref{sec:related_work} summarizes the existing literature on event-based VO. Sec. \ref{sec:methodology} presents the complete pipeline of Gaussian-process continuous-time VO. Sec. \ref{sec:experiments} evaluates the system and ESVO on the MVSEC dataset and discusses the results. Sec. \ref{sec:conclusion} presents the summary of the work.

\begin{figure*}
  \centering
  \includegraphics[width=1\linewidth]{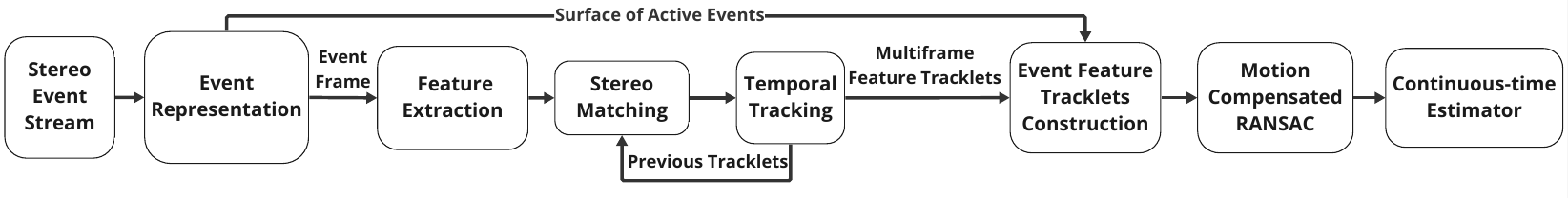}
  \caption{\footnotesize{Overview of the presented VO pipeline. The system takes an asynchronous stereo event stream and clusters the events into event frames and SAEs. Features are detected and tracked in the event frames and each feature is assigned an event time from the SAE. The resulting asynchronous event feature tracklets are filtered with a motion-compensated RANSAC that accounts for their asynchronous times. This gives a consistent inlier tracklet set and motion prior for the continuous-time estimator.}}
  \label{fig:eVO_pipeline}
\end{figure*}

\section{RELATED WORK}
\label{sec:related_work}
Event-based motion estimation techniques can be described by their handling of asynchronous event times as either grouping times into discrete frames (Sec. \ref{sec:lit_e_discrete_VO}) or considering times individually (Sec. \ref{sec:lit_e_continuous_VO}).

\subsection{Grouped-time Approaches} \label{sec:lit_e_discrete_VO}
Grouping event data together creates data frames like traditional cameras. Traditional VO pipelines can then be used to estimate camera poses at the discrete times  assigned to the event frames. These frame times reduce the temporal resolution of the data by replacing all the individual events in a frame with a single time.

Initial event-based VO research focuses on simplified scenarios, e.g., 2D planar motion \cite{2D_eSLAM} or  rotation-only motion \cite{Kim2014_eMosaic_Track}. 
Research extends estimation to $SE(3)$ motion by incorporating complementary sensors into the estimation pipeline. Kueng et al. \cite{Kueng2016_tracking_with_frame_event} fuse both events and traditional-camera frames to detect features on the image frames and track them using events. Weikersdorfer et al. \cite{eSLAM_withRGBD} use both event and RGB-D cameras to provide depth information for each event and create a 3D map for localization.

Event-based Visual Inertial Odometry (VIO) systems specifically include inertial measurement units (IMUs) and are a popular area of research. Zhu et al. \cite{EVIO} and  Rebecq et al. \cite{Rebecq:VIOKeyFrame} accumulate events in a spatial-temporal window to reconstruct frames for feature tracking. The resulting tracklets are used to minimize reprojection and initial error for pose estimation. Ultimate-SLAM \cite{Ultimate_SLAM} extends \cite{Rebecq:VIOKeyFrame} to use the IMU to generate motion-compensated event frames and reformulates the cost function for camera egomotion estimation. Chen et al. \cite{ESVIO} extract features using an asynchronous feature detector (Arc* \cite{Arc*})  from a stereo pair of event cameras and adopt an estimation pipeline similar to \cite{Ultimate_SLAM}. IMU Dynamic Vision Sensor Odometry using Lines (IDOL) \cite{IDOL} uses an alternative VIO paradigm. It uses Gaussian process regression to preintegrate IMU measurements and associate each event with timely accurate IMU data. This maintains the temporal resolution of the event data but still estimates the trajectory at the discrete states of the event frames. 
\squeezeWords

Kim et al. \cite{3Filter_estimation} estimate $SE(3)$ motion using only event-based cameras. They interleave three filters to estimate camera motion, intensity-based frames and scene depth. EVO \cite{EVO} improves computational performance by interleaving geometric semidense mapping \cite{EMVS} and image-to-model alignment for pose estimation. ESVO \cite{ESVO} uses parallel tracking and mapping to estimate the egomotion trajectory and a semidense 3D scene reconstruction. Hadviger et al. \cite{feature_based_ESVO} present a feature-based stereo VO pipeline using events, which group events into frames and then adopts a similar estimation framework to a traditional frame-based stereo VO pipeline \cite{frame_based_SVO}.
\squeezeWords

Discrete-time event-based VO groups multiple event times into a single time. This is helpful for feature detection and tracking but approximates the time of individual event features and reduces the temporal resolution of the measured data. This approximation is inconsistent with the asynchronous nature of event cameras and introduces potential measurement errors.
This paper instead presents a VO pipeline that estimates the camera trajectory from individual event times and maintains the temporal resolution of event cameras. It uses only an event-camera stereo pair and can be implemented with either frame-based or event-based feature detection and tracking.

\subsection{Individual-time Approaches}\label{sec:lit_e_continuous_VO}
Including all the individual, possibly unique event times in the estimation maintains the temporal resolution of event cameras but defines an underconstrained problem.  Similar estimation problems are solved for rolling-shutter cameras and scanning lidars using continuous-time estimation \cite{STEAM, furgale2015continuous}. These techniques estimate the camera trajectory as a continuous function where pose can be queried at any time in the estimation window. A comparison of discrete and continuous-time trajectories can be found in \cite{CT_vs_DT}.

Mueggler et al. \cite{CT-eVIO_TRO} use a continuous-time pose estimation framework that uses IMU measurements and represents the trajectory as cumulative cubic B-splines. This avoids grouping event feature times and maintains their temporal resolution, but requires preprocessing to obtain a scene map.

Wang et al. \cite{wang2022_CT_eVO} use volumetric contrast maximization for continuous-time estimation. The trajectory is initialized with an Ackermann motion model \cite{Ackermann_motion_model}, and globally optimized with a B-spline-based continuous-time estimation framework. This allows the estimator to maintain the native temporal resolution of the event data but limits it to planar motion. %

Liu et al. \cite{liu2022_CT_eVO} also estimate continuous-time trajectories with Gaussian process regression and a WNOA motion prior, but in a monocular VO system. Their estimation runs asynchronously and considers individual event times but, unlike traditional VO pipelines, couples tracklet outlier rejection with the motion estimation which may introduce different sources of error. The evaluation of their algorithm on real-world event datasets is also limited to five-second sequences.
\squeezeWords

This paper presents a complete continuous-time event-based stereo VO pipeline that maintains individual event times in the trajectory estimation. In contrast to these existing works, it maintains temporal resolution with either frame-based or event-based feature detection and tracking and with a RANSAC formulation \cite{RANSAC_MC}  that separates outlier rejection from estimation. It uses a WNOA motion prior to estimate a full $SE(3)$ trajectory directly from event tracklets and their unique timestamps. This approach takes full advantage of the asynchronous nature of the event cameras and allows pose to be queried at any time in the estimation window.

\section{METHODOLOGY}
\label{sec:methodology}
This paper presents an event-based stereo VO system that uses the native temporal resolution of event-based cameras for estimation (Fig. \ref{fig:eVO_pipeline}). %
Features are detected by clustering the event streams of each camera while maintaining the unique event times and tracked in a traditional frame-based manner (Sec. \ref{sec:sub_feature}). This allows for the use of any frame-based feature detection and tracking method or could be directly replaced by event-based approaches. The resulting asynchronous tracklets are then filtered for outliers with a motion-compensated RANSAC (Sec. \ref{sec:sub_RANSAC}). This removes outliers more accurately than other methods by accounting for the different tracklet times. The camera trajectory is then estimated from all the unique tracklet states using Gaussian process regression with a WNOA prior (Sec. \ref{sec:sub_optimization}). This results in a continuous-time VO system that estimates the camera pose at each unique tracklet time and can be queried for the pose at any other time in the estimation window.

\subsection{Event Feature Extraction and Matching}
\label{sec:sub_feature}
Geometric features are extracted from clustered events (Sec. \ref{sec:ssub_ecluster}) using traditional frame-based feature detection techniques, and then matched between the stereo pair and through time to construct feature tracklets (Sec. \ref{sec:ssub_feature}). The temporal resolution of event cameras is maintained by assigning (possibly unique) times to each feature from the corresponding event in a Surface of Active Events (SAE) \cite{SAE_reference}. This allows the pipeline to use any frame-based feature detector and tracker and still create asynchronous tracklets for estimation. It could also be directly replaced with event-based methods, e.g., Arc* \cite{Arc*} or HASTE \cite{HASTE}.

\subsubsection{Event Clustering}\label{sec:ssub_ecluster}
The stereo event stream is rectified and clustered to construct new SAEs and new binary event frames (Fig. \ref{fig:eVO_pipeline}). 
The left and right event streams are synchronously clustered by registering events within a user-specified time window or until either camera registers more than a user-specified number of events in that time. These thresholds define a minimum effective frame rate when the scene changes slowly (e.g., small motion) and a faster frame rate when it changes rapidly (e.g., high-speed motion), while keeping the left and right frames synchronized for feature matching.
The SAE records the most recent event timestamp of each pixel location and is used to maintain the asynchronous nature of the event camera.
The binary event frame is a grey image where white pixels denote events, regardless of polarity.

\subsubsection{Feature Detection and Matching}\label{sec:ssub_feature}
Features are independently detected in the left and right binary event frames for each stereo pair of clusters (Fig. \ref{fig:eVO_pipeline}). These features are then matched using a quad-matching scheme. Features in the current left frame are first matched to the current right frame. The matched features are then successively matched from the current right frame to the previous right frame, from the previous right frame to previous left frame and finally from the previous left frame back to the current left frame. Features successfully matched to all these pairs are kept as tracklets. This feature detection and tracking can use any traditional frame-based approach.

The timestamp of each tracklet state is assigned from the nearest event in the associated SAE. This allows traditional frame-based feature detection methods to detect asynchronous tracklets that maintain the original temporal resolution of the event data. Event-based feature detection and tracking algorithms generate these asynchronous tracklets directly and could be used instead.

Tracklets are extended beyond consecutive frames by matching the new tracklets to previously detected tracklets in earlier frames. This is done by independently checking each new tracklet against a user-specified number of previous frames and recording any additional matches.

The tracklets are filtered using user-specified thresholds to discard those that have \begin{inparaenum} \item a large time difference between stereo features, \item a small disparity between stereo features, \item a short length, and \item a short time\end{inparaenum}.  These filtering schemes remove incorrect matches and improve tracklet quality. This feature extraction and matching process is performed independently at different image resolutions to detect and track features of different size. The resulting tracklets are then processed together to remove outliers.

\subsection{Outlier Rejection}
\label{sec:sub_RANSAC}
Traditional VO pipeline uses Random Sample Consensus (RANSAC) \cite{VO+RANSAC} to remove tracklet outliers before estimation. Traditional RANSAC assumes tracklet states occur at common times and uses a discrete transform motion model. This assumption is incorrect for asynchronous event tracklets and this paper instead uses MC-RANSAC \cite{RANSAC_MC}, which makes no  assumptions about common state times and uses a constant-velocity model in $SE(3)$. The fast version of MC-RANSAC (Sec. \ref{sec:ssub_fast_MCRANSAC}) is used to find an initial inlier set by repeatedly selecting tracklets, calculating the constant-velocity model, and segmenting the tracklets into inliers and outliers based on a user-specified threshold. This process is repeated a user-specified number of times and then the largest inlier set is refined using the full iterative MC-RANSAC (Sec. \ref{sec:ssub_iter_MCRANSAC}) to find the final inlier and outlier segmentation (Fig. \ref{fig:eVO_pipeline}). 
Both versions of MC-RANSAC are compared to traditional RANSAC in \cite{RANSAC_MC}.%

\subsubsection{Fast MC-RANSAC}\label{sec:ssub_fast_MCRANSAC}
The set of tracklets between two event clusters is a number, $M_{\text{clst}}$, of stereo measurements of different landmarks at possibly unique times. MC-RANSAC segments these tracklets into inliers and outliers by finding the most tracklets that can be explained by a single velocity. The constant $SE(3)$ velocity of the sensor, $\boldsymbol{\varpi} \in \mathbb{R}^{6 \times 1}$, is 
\begin{equation*}
\boldsymbol{\varpi} = \begin{bmatrix} \mathbf{v} \\ \boldsymbol{\omega}\end{bmatrix}, 
\end{equation*}
where $\mathbf{v}, \boldsymbol{\omega} \in \mathbb{R}^{3 \times 1}$ are linear and angular velocity components, respectively.
This constant velocity over a time, $\Delta t$, gives the relative $SE(3)$ transformation, 
\begin{equation*}
\mathbf{T} = \exp{(\Delta t \boldsymbol{\varpi}^{\wedge})},
\end{equation*}
where $\exp(\cdot)$ is the matrix exponential and $(\cdot)^{\wedge}$ is the lifting operator that converts $\mathbb{R}^{6 \times 1}$ to $\mathbb{R}^{4 \times 4}$ \cite{TB_book2017}.

A velocity can be calculated from tracklets faster but less accurately by minimizing the error in Euclidean space, 
\begin{equation}\label{eq:fast_cost_func}
J_{\text{fast}}(\boldsymbol{\varpi}) = \frac{1}{2} \sum_{j=1}^{M_{\text{rand}}} \mathbf{e}_{\text{fast},j,k}^T\mathbf{e}_{\text{fast},j,k},
\end{equation}
where $M_{\text{rand}} \leq M_{\text{clst}}$ is the the number of randomly selected tracklets and $\mathbf{e}_{\text{fast},j,k} \in \mathbb{R}^{3 \times 1}$ is the motion-model error,
\begin{equation}\label{eq:eq:fast_err}
\mathbf{e}_{\text{fast},j,k} = \mathbf{P} (\mathbf{p}_k^{j,k} - \mathbf{z}(\mathbf{T}_{k,k'}, \mathbf{p}_{k'}^{j,k'})),
\end{equation}
where $\mathbf{P}$ is the mapping from homogeneous to 3D coordinates, $\mathbf{p}_k^{j,k} \in \mathbb{R}^{4 \times 1}$ is the position of the $j^{\text{th}}$ landmark relative to the camera pose at time $t_k$, $\mathbf{T}_{k,k'}$ is the transformation from timestamp $t_{k'}$ to $t_k$, $t_{k'}< t_k$ is the earlier time of the tracklet segment, and $\mathbf{z}(\cdot)$ is the constant-velocity motion model, 
\begin{equation*}
\mathbf{z}(\mathbf{T}_{k, k'},\mathbf{p}^{j, k'}_{k'} )  = \mathbf{T}_{k, k'}\mathbf{p}^{j, k'}_{k'} 
\end{equation*}

Assuming a short duration tracklet, the corresponding small transformation is approximated as
\begin{equation}\label{eq:approx_T}
\mathbf{T} \approx \mathbf{1} + \boldsymbol{\xi}^{\wedge} = \mathbf{1} + \Delta t \boldsymbol{\varpi}^{\wedge},
\end{equation}
where $\mathbf{1} \in \mathbb{R}^{4 \times 4}$ is the identity matrix, $\boldsymbol{\xi}\in \mathbb{R}^{6 \times 1}$ is the pose in vector form and $\Delta t$ is the duration of the tracklet.
Substituting \eqref{eq:approx_T} into \eqref{eq:eq:fast_err} approximates the error term as,
\begin{equation}\label{eq:fast_err_new}
\begin{aligned}
\mathbf{e}_{\text{fast},j,k} & \approx \mathbf{P}(\mathbf{p}_{k}^{j,k} - (\mathbf{1} + \Delta t_{k,k'}\boldsymbol{\varpi}^{\wedge})\mathbf{p}_{k'}^{j,k'}) \\
& = \mathbf{d}_{j,k} - \Delta t_{k,k'} \mathbf{D}_{j,k}\boldsymbol{\varpi},
\end{aligned}
\end{equation}
where 
\begin{equation*}
\begin{aligned}
& \Delta t_{k,k'} = t_{k} - t_{k'}, \\
&\mathbf{d}_{j,k}  = \mathbf{P}(\mathbf{p}_{k}^{j,k} - \mathbf{p}_{k'}^{j,k'}),\\
&\mathbf{D}_{j,k} = \mathbf{P}(\mathbf{p}_{k'}^{j,k'})^{\odot},
\end{aligned}
\end{equation*}
and $(\cdot)^{\odot}$ is the $\mathbb{R}^{4\times 1}$ to $\mathbb{R}^{4\times 6}$ operator \cite{TB_book2017}.
Substituting \eqref{eq:fast_err_new} into the Euclidean space cost function in \eqref{eq:fast_cost_func}, differentiating with respect to $\boldsymbol{\varpi}$, and setting the result equal to zero gives the velocity best describing a set of  $M_{\text{rand}}$ tracklets,
\begin{equation*}
\boldsymbol{\varpi}
 = \left(\sum_{j=1}^{M_{\text{rand}}} \Delta t_{j,k}^2 \mathbf{D}_{j,k}^T\mathbf{D}_{j,k}\right)^{-1}\left(\sum_{j=1}^{M_{\text{rand}}} \Delta t_{j,k} \mathbf{D}_{j,k}^T\mathbf{d}_{j,k}\right).
\end{equation*}
This velocity can then be used to segment all $M_{\text{clst}}$ tracklets into inliers and outliers by comparing their reprojection error relative to tracklet length,
\begin{equation}\label{eq:err_ransac_classify}
e_{\text{rel}} = \frac{\mathbf{y}_{j,k} - \mathbf{s}(\mathbf{z}(\exp(\Delta t_{k,k'} \boldsymbol{\varpi}^{\wedge}),  \mathbf{p}_{k'}^{j,k'}))}{\| \mathbf{y}_{j,k} - \mathbf{y}_{j,k'}\|},
\end{equation}
to a user-specified threshold, where $\mathbf{y}_{j,k}$ is a measurement of the $j^{\text{th}}$ landmark at time $t_k$ and $\mathbf{s}(\cdot)$ is the nonlinear camera projection model from a 3D landmark to a 2D image point.

The process of randomly selecting a small number of tracklets, quickly calculating a velocity from them, and then using this velocity to classify all the tracklets as inliers or outliers is repeated a user-specified number of times. The largest inlier set found is used as the initial segmentation for the full iterative MC-RANSAC.

\subsubsection{Iterative MC-RANSAC}\label{sec:ssub_iter_MCRANSAC}
A more accurate iterative MC-RANSAC approach minimizes the reprojection error of each tracklet in image space with the cost function,
\begin{equation}\label{eq:iter_cost_func}
J_{\text{iter}}(\boldsymbol{\varpi}) = \frac{1}{2} \sum_{j=1}^{M_\text{fast}}\mathbf{e}_{\text{iter},j,k}^T \mathbf{R}^{-1}_{j,k}\mathbf{e}_{\text{iter},j,k},
\end{equation}
where $\mathbf{R}_{j,k}$ is the covariance matrix for the measurements of the $j^\text{th}$ tracklet and $M_{\text{fast}}$  is the number of inliers found by the fast MC-RANSAC. The reprojection error is
\begin{equation}\label{eq:iter_err}
\mathbf{e}_{\text{iter, j,k}} = \mathbf{y}_{j,k} - \mathbf{s}(\mathbf{z}(\exp{(\Delta t_{k,k'} \boldsymbol{\varpi}^{\wedge})},\mathbf{p}^{j, k'}_{k'} )).
\end{equation}

The error function is linearized by representing the velocity as a nominal value, $\bar{\boldsymbol{\varpi}}$, and a small perturbation, $\delta \boldsymbol{\varpi}$,
\begin{equation}\label{eq:iter_linearized_error}
\boldsymbol{\varpi} = \bar{\boldsymbol{\varpi}} + \delta \boldsymbol{\varpi}.
\end{equation}
Substituting \eqref{eq:iter_err} and \eqref{eq:iter_linearized_error} into \eqref{eq:iter_cost_func}, differentiating with respect to $\delta \boldsymbol{\varpi}$, and setting the result equal to zero gives the perturbation that minimizes the linearization, 
\begin{equation*}
\delta \boldsymbol\varpi^* = \left(\sum_{j=1}^{M_\text{fast}} \mathbf{H}_{j,k}^{T}\mathbf{R}_{j,k}^{-1}\mathbf{H}_{j,k}\right)^{-1}\left(\sum_{j=1}^{M_\text{fast}} \mathbf{H}_{j,k}^{T}\mathbf{R}_{j,k}^{-1}\mathbf{e}_{\text{iter},j,k}\right), 
\end{equation*}
where $\mathbf{H}_{j,k}$ is the Jacobian of error function in \eqref{eq:iter_err},
\begin{equation*}
\mathbf{H}_{j,k} = \left .\frac{\partial \mathbf{s}}{\partial \mathbf{z}} \right |_{\bar{\mathbf{z}}} \Delta t_{k,k'} \mathbf{T}_{k,k'} (\mathbf{p}_{k'}^{j,k'})^{\odot} \boldsymbol{\mathcal{T}}_{k,k'}^{-1}\boldsymbol{\mathcal{J}}_{k,k'},
\end{equation*}
where the partial derivative of the sensor model is evaluated at the nominal value, $\boldsymbol{\mathcal{T}}_{k,k'}$ is the adjoint of $SE(3)$ and $\boldsymbol{\mathcal{J}}_{k,k'}$ is the left Jacobian of $SE(3)$ \cite{TB_book2017}.

This process is iterated until convergence to find the velocity best describing the initial inlier set found by fast MC-RANSAC, 
\begin{equation*}
\bar{\boldsymbol{\varpi}} \gets \bar{\boldsymbol{\varpi}} + \delta \boldsymbol\varpi^*.
\end{equation*}
This velocity is then used to segment all $M_{\text{clst}}$ tracklets into inliers and outliers by comparing their reprojection error using \eqref{eq:err_ransac_classify} to a user-specified threshold. This is more accurate outlier rejection than using fast MC-RANSAC alone.

\subsection{Continuous-time Trajectory Optimization}
\label{sec:sub_optimization}
Traditional discrete-time estimation requires at least three measurements at every estimation state. This is often achieved in event-based VO by grouping event features to common times, which reduces the temporal resolution of event cameras. Continuous-time trajectory estimation can instead operate directly on asynchronous tracklets, which may result in estimation states with less than three measurements, by incorporating a motion prior or basis function.
This allows continuous-time estimation techniques to maintain the temporal resolution of the event cameras.

The inlier tracklets from MC-RANSAC are used to define the trajectory optimization problem (Fig. \ref{fig:eVO_pipeline}). Each unique tracklet timestamp defines a state in the estimation problem. The continuous-time trajectory is estimated from these states using Gaussian process regression with a WNOA prior \cite{STEAM} (Sec. \ref{sec:ssub_Est}). This iterative process uses the velocities found during MC-RANSAC as an initial condition.

The WNOA motion prior is physically founded and accounts for real-world kinematics, unlike other continuous-time parametrizations that enforce mathematical smoothness independent of its physical plausibility. It is compared quantitatively to other estimation techniques in \cite{STEAM, TongIJRR2013}. The resulting trajectory can be queried for the camera pose at any timestamp in the estimation window (Sec. \ref{sec:ssub_CT_query}).

\subsubsection{WNOA Estimator}\label{sec:ssub_Est}
The estimated states are defined as
\begin{equation*}\label{eq:traj_state}
\mathbf{x} =  \{\mathbf{T}_{k,1}, \boldsymbol{\varpi}_k , \mathbf{p}_1^{j,1}\}_{j=1,\cdots,M , k=1,\cdots,K},
\end{equation*}
where $\mathbf{T}_{k,1} \in SE(3)$ is the pose at the time $t_k$ relative to the initial pose, $\boldsymbol{\varpi}_k$ is the corresponding 6DOF body-centric velocity and $\mathbf{p}_1^{j,1}$ is the position of the $j^{\text{th}}$ landmark relative to the initial pose. The camera trajectory is represented by discrete estimated states, $\mathbf{T}_{k,1}$ and $\boldsymbol{\varpi}_k$, which can be denoted with a slight abuse of notation as $\mathbf{x}_k = \{\mathbf{T}_{k,1}, \boldsymbol{\varpi}_k\}$. The acceleration, $\dot{\boldsymbol{\varpi}}$, is assumed to be a zero-mean, white-noise Gaussian process,
\begin{equation}\label{eq:WNOA_accel}
\dot{\boldsymbol{\varpi}} \sim \mathcal{GP}(\mathbf{0}, \mathbf{Q}_c \delta(t-t^{\prime})),
\end{equation}
where $\mathbf{Q}_c \in \mathbb{R}^{6 \times 6}$ is a diagonal power spectral density matrix, and $\delta(\cdot)$ is the Dirac delta function.

The WNOA assumption defines a locally constant velocity motion model. A local trajectory state, $\boldsymbol{\gamma}_k$, can be defined as a continuous-time function with respect to the global trajectory state,
\squeezeWords
\begin{equation*}
\boldsymbol{\gamma}_k(\tau) = \begin{bmatrix}\boldsymbol{\xi}_k(\tau) \\ \dot{\boldsymbol{\xi}_k}(\tau)\end{bmatrix} = \begin{bmatrix}
\ln(\mathbf{T}_k(\tau)\mathbf{T}_{k}^{-1})^{\vee} \\
\boldsymbol{\mathcal{J}}(\ln(\mathbf{T}_k(\tau)\mathbf{T}_{k}^{-1})^{\vee})^{-1}\boldsymbol{\varpi}_{k}(\tau) \end{bmatrix},
\end{equation*}
where $\mathbf{T}_k(\tau)$ is the pose at  time $t_k \leq \tau \leq t_{k+1}$, $\boldsymbol{\mathcal{J}}(\cdot)^{-1}$ is the inverse left Jacobian function, $\ln(\cdot)$ is the inverse exponential map, and  $(\cdot)^{\vee}$ is the inverse lifting operator \cite{TB_book2017}.

The estimator minimizes a joint cost function, 
\begin{equation}\label{eq:joint_cost}
J_{\text{joint}}(\mathbf{x}) = J_{\text{prior}}(\mathbf{x}) + J_{\text{meas}}(\mathbf{x}),
\end{equation}
where $J_{\text{prior}}(\mathbf{x})$ is the motion prior cost function and $J_{\text{meas}}(\mathbf{x})$ is the measurement cost term. The prior cost function penalizes trajectory states that deviate from WNOA assumption. The prior cost function is
\begin{equation*}
J_{\text{prior}}(\mathbf{x}) = \frac{1}{2} \sum_{k=1}^{K-1}\mathbf{e}_{\text{prior},k+1,k}^T\mathbf{Q}^{-1}_k(t_{k+1})\mathbf{e}_{\text{prior},k+1,k},
\end{equation*}
where the prior error is
\begin{equation}\label{eq:prior_err}
\mathbf{e}_{\text{prior},k+1,k} = \begin{bmatrix}
\ln(\mathbf{T}_{k+1,1}\mathbf{T}_{k,1}^{-1})^{\vee} - (t_{k+1}-t_{k})\boldsymbol{\varpi}_{k} \\
\boldsymbol{\mathcal{J}}(\ln(\mathbf{T}_{k+1,1}\mathbf{T}_{k,1}^{-1})^{\vee})^{-1}\boldsymbol{\varpi}_{k+1} - \boldsymbol{\varpi}_{k},
\end{bmatrix},
\end{equation}
and the prior covariance matrix at $t_{\tau}$ from $t_{k}$ is,
\begin{equation}\label{eq:prior_cov}
\mathbf{Q}_k(\tau) = \begin{bmatrix}
\frac{1}{3}\Delta t_{\tau,k}^3 \mathbf{Q}_c & \frac{1}{2} \Delta t_{\tau,k}^2 \mathbf{Q}_c \\
\frac{1}{2} \Delta t_{\tau,k}^2 \mathbf{Q}_c & \Delta t_{\tau,k} \mathbf{Q}_c
\end{bmatrix},
\end{equation}
where $\Delta t_{\tau, k} = \tau - t_k$ and  $\mathbf{Q}_c \in \mathbb{R}^{6 \times 6}$ is the power spectral density matrix defined in \eqref{eq:WNOA_accel}.

The measurement cost function that minimizes the feature tracklet reprojection error is
\begin{equation*}
J_{\text{meas}}(\mathbf{x})= \frac{1}{2} \sum_{j,k}\mathbf{e}_{\text{meas},j,k}^T \mathbf{R}_{j,k}^{-1} \mathbf{e}_{\text{meas},j,k},
\end{equation*}
where $\mathbf{R}_{j,k} \in \mathbb{R}^{3 \times 3}$ is the measurement covariance matrix of the $j^{\text{th}}$ landmark viewed from the $
k^{\text{th}}$ state and the tracklet reprojection error is
\begin{equation}\label{eq:meas_err}
\mathbf{e}_{\text{meas},j,k} = \mathbf{y}_{j,k} - \mathbf{s}(\mathbf{z}(\mathbf{T}_{k,1}, \mathbf{p}_1^{j,1})).
\end{equation}

The optimal camera trajectory is obtained by optimizing the joint cost function,
\begin{equation*}\label{eq:argmin_state}
\mathbf{x}^* =\operatorname*{arg\,min}_{\mathbf{x}} \{J_{\text{meas}}(\mathbf{x}) + J_{\text{prior}}(\mathbf{x})\},
\end{equation*}
using the Gauss-Newton method. The states are approximated as operating points, $\mathbf{x}_{\text{op}}= \{\mathbf{T}_{\text{op}}, \boldsymbol{\varpi}_{\text{op}}, \mathbf{p}_{\text{op}}\}$,  and perturbations, $\delta \mathbf{x} = \{\delta\boldsymbol{\xi}, \delta\boldsymbol{\varpi}, \delta\boldsymbol{\zeta}\}$, linearizing \eqref{eq:joint_cost} as
\begin{equation}\label{eq:linearized_joint}
J_{\text{joint}}(\mathbf{x}) = J_{\text{joint}}(\mathbf{x}_{\text{op}}) - \mathbf{b}^T\delta\mathbf{x} + \frac{1}{2}\delta\mathbf{x}^T\mathbf{A}\delta\mathbf{x},
\end{equation}
where
\begin{equation*}
\begin{aligned}
\mathbf{A} & = 
\sum_{j,k} \mathbf{P}_{j,k}^T\mathbf{G}_{j,k}^T\mathbf{R}_{j,k}^{-1}\mathbf{G}_{j,k}\mathbf{P}_{j,k}\\ 
& \qquad + \sum_k \mathbf{P}_{k}^T\mathbf{E}_{k}^T\mathbf{Q}_k^{-1}(t_{k+1})\mathbf{E}_{k}\mathbf{P}_{k}, \\
\mathbf{b}  & = \sum_{j,k} \mathbf{P}_{j,k}^T\mathbf{G}_{j,k}^T\mathbf{R}_{j,k}^{-1}\mathbf{e}_{\text{meas},j,k}\\
& \qquad + \sum_k \mathbf{P}_{k}^T\mathbf{E}_{k}^T\mathbf{Q}_k^{-1}(t_{k+1})\mathbf{e}_{\text{prior},k+1,k},
\end{aligned}
\end{equation*}
where  $\mathbf{P}_{j,k}$ and $\mathbf{P}_{k}$ are matrices to pick the specific components of the total perturbation, 
$\mathbf{G}_{j,k}$ is the Jacobian of~\eqref{eq:meas_err},
\squeezeWords
\begin{equation*}
\mathbf{G}_{j,k} = \left .\frac{\partial \mathbf{s}}{\partial \mathbf{z}} \right |_{\bar{\mathbf{z}}}\begin{bmatrix}(\mathbf{T}_{\text{op},k,1}\mathbf{p}_{\text{op},1}^{j,1})^{\odot} & \mathbf{0} & \mathbf{T}_{\text{op},k,1} \begin{bmatrix} \mathbf{1} \\ \mathbf{0}^T\end{bmatrix} \end{bmatrix},
\end{equation*}
and $\mathbf{E}_{k}$ is the Jacobian of the prior error function in \eqref{eq:prior_err},
\begin{equation*}
\mathbf{E}_k = \begin{bmatrix}\mathbf{E}_{11} & \Delta t_{k+1,k}\mathbf{1} & \mathbf{E}_{13} & \mathbf{0}\\
\mathbf{E}_{21} & \mathbf{1} & \mathbf{E}_{23} & \mathbf{E}_{24}  \end{bmatrix},\\
\end{equation*}
where
\begin{equation*}
\begin{aligned}
& \mathbf{E}_{11} = \boldsymbol{\mathcal{J}}_{k+1,k}^{-1}\boldsymbol{\mathcal{T}}_{k+1,k}, 
& \mathbf{E}_{13}  =  \mathbf{E}_{24} = -\boldsymbol{\mathcal{J}}_{k+1,k}^{-1} , \\
& \mathbf{E}_{21} = \frac{1}{2} \boldsymbol{\varpi}^{\curlywedge}_{k+1}\boldsymbol{\mathcal{J}}_{k+1,k}^{-1}\boldsymbol{\mathcal{T}}_{k+1,k},
& \mathbf{E}_{23} = -\frac{1}{2} \boldsymbol{\varpi}^{\curlywedge}_{k+1}\boldsymbol{\mathcal{J}}_{k+1,k}^{-1} , 
\end{aligned}
\end{equation*}
and $(\cdot)^{\curlywedge}$ is the $\mathbb{R}^{6\times 1}$ to $\mathbb{R}^{6 \times 6}$ operator \cite{TB_book2017}.

Taking the derivative of \eqref{eq:linearized_joint} with respect to $\delta \mathbf{x}^*$, setting the result equal to zero and solving the resulting linear system, $\mathbf{A}\delta \mathbf{x}^* = \mathbf{b}$, gives the perturbation that minimizes the linearization. The estimation states are updated, 
\begin{equation*}
\begin{aligned}
&\mathbf{T}_{\text{op},k,1} \gets \exp(\delta \boldsymbol{\xi}^*)\mathbf{T}_{\text{op},k,1},\\
&\boldsymbol{\varpi}_{\text{op},k}  \gets \boldsymbol{\varpi}_{\text{op},k} + \delta \boldsymbol{\varpi}^*,  \\
&\mathbf{p}_{\text{op},j}  \gets \mathbf{p}_{\text{op},j} + \delta\boldsymbol{\zeta}^*.
\end{aligned}
\end{equation*}
and the process is iterated until convergence.
The final estimate is the landmark position and the continuous-time trajectory represented as discrete state poses, discrete local velocities, and the WNOA prior.

\subsubsection{Querying the continuous-time trajectory}\label{sec:ssub_CT_query}
The continuous-time trajectory can be queried for the camera pose at any time during the estimation window, $t_1 \leq \tau \leq t_K$. The pose is interpolated between the states at the two closest times, $t_m \leq \tau \leq t_n$, using the WNOA prior.
The local trajectory state, $\boldsymbol{\gamma}_m(\tau)$, is interpolated \cite{STEAM}~as
\begin{equation*}
\boldsymbol{\gamma}_{m}(\tau) = \boldsymbol{\Lambda}(\tau)\boldsymbol{\gamma}_{m}(t_m)+ \boldsymbol{\Omega}(\tau)\boldsymbol{\gamma}_{n}(t_n),
\end{equation*}
where
\begin{equation*}
\begin{aligned}
&\boldsymbol{\Lambda}(\tau) = \boldsymbol{\Phi}(\tau, t_m) -  \boldsymbol{\Omega}(\tau)\boldsymbol{\Phi}(t_n, t_m) \\
 &\boldsymbol{\Omega}(\tau) = \mathbf{Q}_m(\tau) \boldsymbol{\Phi}(t_n, \tau)^T \mathbf{Q}_m(t_n)^{-1} \\
&\boldsymbol{\Phi}(\tau, t_m) = \begin{bmatrix} \mathbf{1} & (\tau - t_m)\mathbf{1}\\ \mathbf{0} & \mathbf{1}\end{bmatrix},
\end{aligned}
\end{equation*}
and $\mathbf{Q}_m(\cdot)$ is defined in \eqref{eq:prior_cov}.

This interpolation can also be used to define the estimation states at a subset of the measurement times \cite{STEAM, TongIJRR2013}.

\begin{table*}[tbp]
\scriptsize
\centering
\caption{\footnotesize{The global error of the presented Gaussian process continuous-time approach (GPCT) and ESVO.}}
\begin{tabular}{lccccccccccc} %
\toprule

\multirow{2}{0.3cm}{} & \multirow{2}{0.4cm}{}&  \multicolumn{5}{c}{\textit{Indoor1}} & \multicolumn{5}{c}{\textit{{Indoor3}}}   \\

\cmidrule(lr){3-7}\cmidrule(lr){8-12} 

\multicolumn{1}{c}{}& \multicolumn{1}{c}{}& \multicolumn{1}{c}{$\text{Max}$} & \multicolumn{1}{c}{$\text{Max \%}$} & \multicolumn{1}{c}{$\text{Final \%}$}  & \multicolumn{1}{c}{$\text{RMS}$} & \multicolumn{1}{c}{$\text{St. Dev.}$} & \multicolumn{1}{c}{$\text{Max}$} & \multicolumn{1}{c}{$\text{Max \%}$} & \multicolumn{1}{c}{$\text{Final \%}$} &  \multicolumn{1}{c}{$\text{RMS}$} & \multicolumn{1}{c}{$\text{St. Dev.}$} \\

\cmidrule(lr){3-12}

\multirow{3}{0.4cm}{GPCT} &\multicolumn{1}{c}{tran.} &  \multicolumn{1}{c}{$\mathbf{0.635}$} &  \multicolumn{1}{c}{$\mathbf{7.58\boldsymbol{\%}}$} &  \multicolumn{1}{c}{$\mathbf{5.73\boldsymbol{\%}}$} &  \multicolumn{1}{c}{$\mathbf{0.372}$}  & \multicolumn{1}{c}{$0.192$} &  \multicolumn{1}{c}{$\mathbf{0.306}$} &  \multicolumn{1}{c}{$\mathbf{5.11\boldsymbol{\%}}$} &  \multicolumn{1}{c}{$5.08\%$}  &\multicolumn{1}{c}{$\mathbf{0.218}$} & \multicolumn{1}{c}{$\mathbf{0.059}$}\\ [2pt]

& \multicolumn{1}{c}{rota.} &  \multicolumn{1}{c}{$0.195$} &  \multicolumn{1}{c}{$\mathbf{7.98\boldsymbol{\%}}$} &  \multicolumn{1}{c}{$7.6\%$} &  \multicolumn{1}{c}{$\mathbf{0.087}$}  & \multicolumn{1}{c}{$0.05$} &  \multicolumn{1}{c}{$\mathbf{0.088}$} &  \multicolumn{1}{c}{$\mathbf{4.23\boldsymbol{\%}}$} &  \multicolumn{1}{c}{$4.13\%$}  &\multicolumn{1}{c}{$0.051$} & \multicolumn{1}{c}{$0.023$}\\ [2pt]

& \multicolumn{1}{c}{$SE(3)$} &  \multicolumn{1}{c}{$\mathbf{0.639}$} &  \multicolumn{1}{c}{$\mathbf{7.18\boldsymbol{\%}}$} &  \multicolumn{1}{c}{$\mathbf{5.78\boldsymbol{\%}}$} &  \multicolumn{1}{c}{$\mathbf{0.382}$}  & \multicolumn{1}{c}{$0.196$} &  \multicolumn{1}{c}{$\mathbf{0.319}$} &  \multicolumn{1}{c}{$\mathbf{4.95\boldsymbol{\%}}$} &  \multicolumn{1}{c}{$4.93\%$}  &\multicolumn{1}{c}{$\mathbf{0.224}$} & \multicolumn{1}{c}{$\mathbf{0.061}$}\\ [2pt]

\multirow{3}{0.4cm}{ESVO}&  \multicolumn{1}{c}{tran.}&  \multicolumn{1}{c}{$0.642$} &  \multicolumn{1}{c}{$7.66\%$} &  \multicolumn{1}{c}{$6.61\%$} &  \multicolumn{1}{c}{$0.477$}  & \multicolumn{1}{c}{$\mathbf{0.152}$} &\multicolumn{1}{c}{$0.683$} &  \multicolumn{1}{c}{$11.4\%$} &  \multicolumn{1}{c}{$\mathbf{4.33\boldsymbol{\%}}$} & \multicolumn{1}{c}{$0.232$} & \multicolumn{1}{c}{$0.104$} \\ [2pt]

&  \multicolumn{1}{c}{rota.}&  \multicolumn{1}{c}{$\mathbf{0.189}$} &  \multicolumn{1}{c}{$8.87\%$} &  \multicolumn{1}{c}{$\mathbf{6.8}\boldsymbol{\%}$} &  \multicolumn{1}{c}{$0.139$}  & \multicolumn{1}{c}{$\mathbf{0.039}$} &\multicolumn{1}{c}{$0.165$} &  \multicolumn{1}{c}{$8.66\%$} &  \multicolumn{1}{c}{$\mathbf{2.39\boldsymbol{\%}}$} & \multicolumn{1}{c}{$\mathbf{0.038}$} & \multicolumn{1}{c}{$\mathbf{0.02}$} \\ [2pt]

&  \multicolumn{1}{c}{$SE(3)$}&  \multicolumn{1}{c}{$0.663$} &  \multicolumn{1}{c}{$7.55\%$} &  \multicolumn{1}{c}{$6.51\%$} &  \multicolumn{1}{c}{$0.497$}  & \multicolumn{1}{c}{$\mathbf{0.155}$} &\multicolumn{1}{c}{$0.697$} &  \multicolumn{1}{c}{$10.9\%$} &  \multicolumn{1}{c}{$\mathbf{4.12\boldsymbol{\%}}$} & \multicolumn{1}{c}{$0.235$} & \multicolumn{1}{c}{$0.103$} \\
\bottomrule
\end{tabular}%
\vspace{-0.5\baselineskip}
\label{tb:stat_GE}
\end{table*}

\begin{table}[tbp]
\centering
\caption{\footnotesize{The relative error of the presented Gaussian process continuous-time approach (GPCT) and ESVO.}}
\label{tb:stat_RE}
\resizebox{0.5\textwidth}{!}{
\begin{tabular}{lccccccc} %
\toprule

\multirow{2}{0.4cm}{}& \multirow{2}{0.4cm}{}&  \multicolumn{3}{c}{\textit{Indoor1}} & \multicolumn{3}{c}{\textit{Indoor3}}    \\

\cmidrule(lr){3-5}\cmidrule(lr){6-8}

\multicolumn{1}{c}{}& \multicolumn{1}{c}{}& \multicolumn{1}{c}{$\text{Max}$}  & \multicolumn{1}{c}{$\text{RMS}$} & \multicolumn{1}{c}{$\text{St. Dev.}$} & \multicolumn{1}{c}{$\text{Max}$}  &  \multicolumn{1}{c}{$\text{RMS}$} & \multicolumn{1}{c}{$\text{St. Dev.}$} \\

\cmidrule(lr){3-8}

\multirow{3}{0.4cm}{GPCT}&  \multicolumn{1}{c}{tran.}  &  \multicolumn{1}{c}{$\mathbf{73\cdot 10^{-3}}$} &  \multicolumn{1}{c}{$\mathbf{7.5\cdot 10^{-3}}$} & \multicolumn{1}{c}{$\mathbf{5.3\cdot 10^{-3}}$}& \multicolumn{1}{c}{$\mathbf{32\cdot 10^{-3}}$}  & \multicolumn{1}{c}{$\mathbf{5.5\cdot 10^{-3}}$} & \multicolumn{1}{c}{$\mathbf{3.6\cdot 10^{-3}}$} \\ [3pt]

&  \multicolumn{1}{c}{rota.}  &  \multicolumn{1}{c}{$\mathbf{18\cdot 10^{-3}}$} &  \multicolumn{1}{c}{$\mathbf{2.4\cdot 10^{-3}}$} & \multicolumn{1}{c}{$\mathbf{1.4\cdot 10^{-3}}$}& \multicolumn{1}{c}{$\mathbf{13\cdot 10^{-3}}$}  & \multicolumn{1}{c}{$\mathbf{2.3\cdot 10^{-3}}$} & \multicolumn{1}{c}{$\mathbf{1.3\cdot 10^{-3}}$} \\ [3pt]

&  \multicolumn{1}{c}{$SE(3)$}  &  \multicolumn{1}{c}{$\mathbf{74\cdot 10^{-3}}$} &  \multicolumn{1}{c}{$\mathbf{7.9\cdot 10^{-3}}$} & \multicolumn{1}{c}{$\mathbf{5.3\cdot 10^{-3}}$}& \multicolumn{1}{c}{$\mathbf{32\cdot 10^{-3}}$}  & \multicolumn{1}{c}{$\mathbf{5.9\cdot 10^{-3}}$} & \multicolumn{1}{c}{$\mathbf{3.6\cdot 10^{-3}}$} \\ [3pt]

\multirow{3}{0.4cm}{ESVO} &  \multicolumn{1}{c}{tran.}  &  \multicolumn{1}{c}{$75\cdot 10^{-3}$} &  \multicolumn{1}{c}{$17\cdot 10^{-3}$} & \multicolumn{1}{c}{$9.2\cdot 10^{-3}$} & \multicolumn{1}{c}{$219\cdot 10^{-3}$} & \multicolumn{1}{c}{$24\cdot 10^{-3}$} & \multicolumn{1}{c}{$19\cdot 10^{-3}$} \\ [3pt]

 &  \multicolumn{1}{c}{rota.}  &  \multicolumn{1}{c}{$43\cdot 10^{-3}$} &  \multicolumn{1}{c}{$7.2\cdot 10^{-3}$} & \multicolumn{1}{c}{$3.9\cdot 10^{-3}$} & \multicolumn{1}{c}{$63\cdot 10^{-3}$} & \multicolumn{1}{c}{$8.3\cdot 10^{-3}$} & \multicolumn{1}{c}{$5.6\cdot 10^{-3}$} \\ [3pt]

  &  \multicolumn{1}{c}{$SE(3)$}  &  \multicolumn{1}{c}{$86\cdot 10^{-3}$} &  \multicolumn{1}{c}{$18\cdot 10^{-3}$} & \multicolumn{1}{c}{$9.2\cdot 10^{-3}$} & \multicolumn{1}{c}{$220\cdot 10^{-3}$} & \multicolumn{1}{c}{$25\cdot 10^{-3}$} & \multicolumn{1}{c}{$19\cdot 10^{-3}$}  \\ 
\bottomrule
\end{tabular}}
\end{table}

\section{EXPERIMENTS}
\label{sec:experiments}
The presented pipeline is evaluated on the MVSEC dataset \cite{MVSEC} and compared against the publicly available ESVO \cite{ESVO}, a discrete event-based stereo VO pipeline. MVSEC consists of complex, nonconstant-velocity motion with stereo event camera data and $100$~Hz ground truth poses for indoor scenes. The \textit{indoor1} and \textit{indoor3} sequences are used since ESVO provides tuning for these sequences. The performance of both algorithms is evaluated using global and relative error (Sec. \ref{sec:sub_metric}) and the results are discussed in Sec. \ref{sec:sub_result}.

A sliding-window version of the system is implemented in MATLAB using LIBVISO2 \cite{LIBVISO2} for feature detection and tracking. The sliding window width is five and LIBVISO2 is run on both full- and half-resolution images. Tracklets are filtered out if they \begin{inparaenum}\item have more than 20ms time difference between stereo features, \item have less than 2px disparity, \item have less than 2px length, and \item last less than 40ms in time\end{inparaenum}. Outlier rejection was done with 10000 iterations of fast MC-RANSAC followed with one call of iterative MC-RANSAC, both using  an inlier threshold of 5\%. The estimator uses the covariances $\mathbf{Q}_c ^{-1}= 50\diag(1 , 1 , 1,10,10,10 )$ and $\mathbf{R}_{j,k}^{-1} = 0.1\diag(5 , 5, 1)$ and terminates when the cost change between two iterations is less than 1\%.

\begin{figure}[tbp]
  \centering%
  \begin{subfigure}{0.49\columnwidth}%
  \centering%
  \includegraphics[height=1.25in]{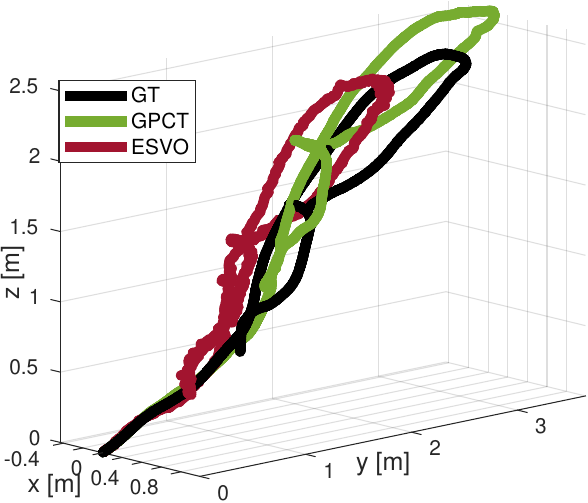}%
  \caption{\footnotesize{MVSEC \textit{indoor1} sequence}}%
  \label{fig:traj_indoor1}%
  \end{subfigure}%
  \begin{subfigure}{0.49\columnwidth}%
   \centering%
  \includegraphics[width=1\linewidth]{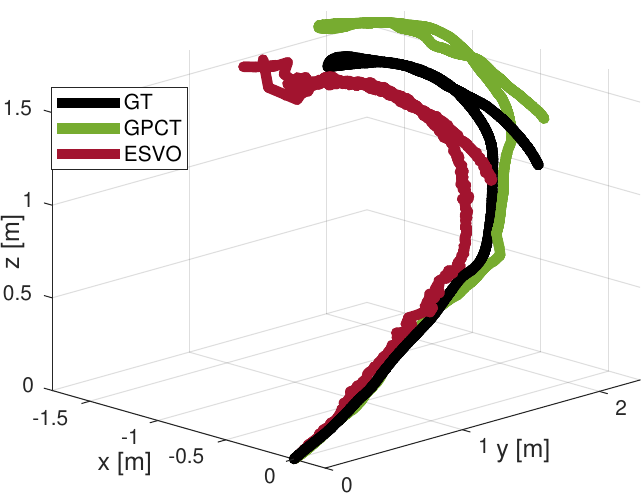}%
  \caption{\footnotesize{MVSEC \textit{indoor3} sequence}}%
  \label{fig:traj_indoor3}%
  \end{subfigure}%
  \caption{\footnotesize{Trajectory plots of the presented Gaussian process continuous-time approach (GPCT), ESVO and ground truth (GT) results in 3D space. GPCT performs better in challenging scenarios like rotation and back-and-forth motions. It also has a smoother trajectory due to the WNOA motion prior.}}%
  \label{fig:traj_plots}%
\end{figure}

\vspace{-0.5ex}
\subsection{Evaluation Metrics}\label{sec:sub_metric}
Global error quantifies the estimator accuracy relative to the initial pose. Relative error quantifies the amount of error in each estimate and is often used to calculate aggregate values over the trajectory. Both can be calculated from ground truth using a single general equation \cite{MVO},
\begin{equation*}
\text{err}(t_m, t_n) = \ln(\mathbf{T}_{m^{\text{GT}},m}\mathbf{T}_{n,m}\mathbf{T}_{m^{\text{GT}},m}^{-1}\mathbf{T}_{n^{\text{GT}},m^{\text{GT}}}^{-1})^{\vee},
\end{equation*}
where $\mathbf{T}_{n,m}$ is the estimated transform to the $n^\text{th}$ frame from the $m^{\text{th}}$ frame, $\mathbf{T}_{n^{\text{GT}},m^{\text{GT}}}$ is the ground truth transform to the $n^\text{th}$ frame from the  $m^{\text{th}}$ frame, and $\mathbf{T}_{m^{\text{GT}},m}$ is the transform to the ground truth $m^{\text{th}}$ frame from the estimate of the $m^{\text{th}}$ frame.
The global error at a time, $t_k$, is then defined relative to the initial pose,
\begin{equation*}\label{eq:GE}
\text{GE}(t_k) = \text{err}(t_k,t_1),
\end{equation*}
and the relative error is defined relative to the previous pose,
\begin{equation*}\label{eq:RE}
\text{RE}(t_k) = \text{err}(t_k,t_{k-1}).
\end{equation*}

The error of the presented continuous-time system is calculated using the timestamps of ESVO. For global error, the continuous-time trajectory is aligned with ESVO's initial pose and then queried at its own state times. For relative error, the continuous-time trajectory is only queried at the discrete state times of ESVO so that both estimators have the same durations between estimates and their relative errors can be compared directly. The high frequency ground truth trajectory is linearly interpolated to the timestamps of the estimated states.

The prototype implementation of the continuous-time system does not run in real time. The system's computational performance should be improved by implementing it in a more efficient language, such as C++, and using keytimes to reduce the number of the estimation states \cite{STEAM, TongIJRR2013}.

\vspace{-0.5ex}
\subsection{Results}\label{sec:sub_result}
The estimated trajectories are evaluated qualitatively relative to the ground truth in 3D (Fig. \ref{fig:traj_plots}). They are evaluated quantitatively by calculating the global and relative errors with respect to ground truth (Fig. \ref{fig:GE_RE}, Tables \ref{tb:stat_GE} and \ref{tb:stat_RE}).
The errors are evaluated statistically using root-mean-squared (RMS), standard deviation (St. Dev.) and maximum error (Max). The maximum global error and final global error are also presented as a percentage of the integration of the norm of the relevant ground-truth component (e.g., path length). The qualitative results demonstrate the benefits of the WNOA prior, with the presented system having a smoother estimated trajectory than that of ESVO. The presented system also quantitatively has smaller RMS relative error and better or equivalent maximum global error than ESVO.

\subsubsection{MVSEC \textit{indoor1}}
The presented system has a similar performance in global error to ESVO (Fig. \ref{fig:GE_indoor1}). The relative error is smaller than ESVO (Fig. \ref{fig:RE_indoor1}) with an RMS value that is two-times better (Table \ref{tb:stat_RE}). This illustrates the better local consistency of the trajectory estimate and explains the smooth trajectory plot in Fig. \ref{fig:traj_indoor1}.

The grey shaded areas in Fig. \ref{fig:GE_indoor1}  denote complex motions where significant error occurs. When the camera undergoes large motions (e.g., large rotation) the observed scene changes drastically. This reduces the quality and quantity of tracklets found by the clustered feature detection and tracking and as a result the quality of the trajectory estimation. This can be improved with better feature tracking, likely specifically designed for event-based cameras.

\subsubsection{MVSEC \textit{indoor3}}
The presented system estimates a smoother trajectory than ESVO in the \textit{indoor3} segment. The estimator qualitatively describes the ground truth motion and is locally consistent. It has almost a four-times better RMS relative error than ESVO (Table \ref{tb:stat_RE})

\begin{figure*} 
  \centering
  \begin{subfigure}{0.49\linewidth}
  \includegraphics[width=1\linewidth]{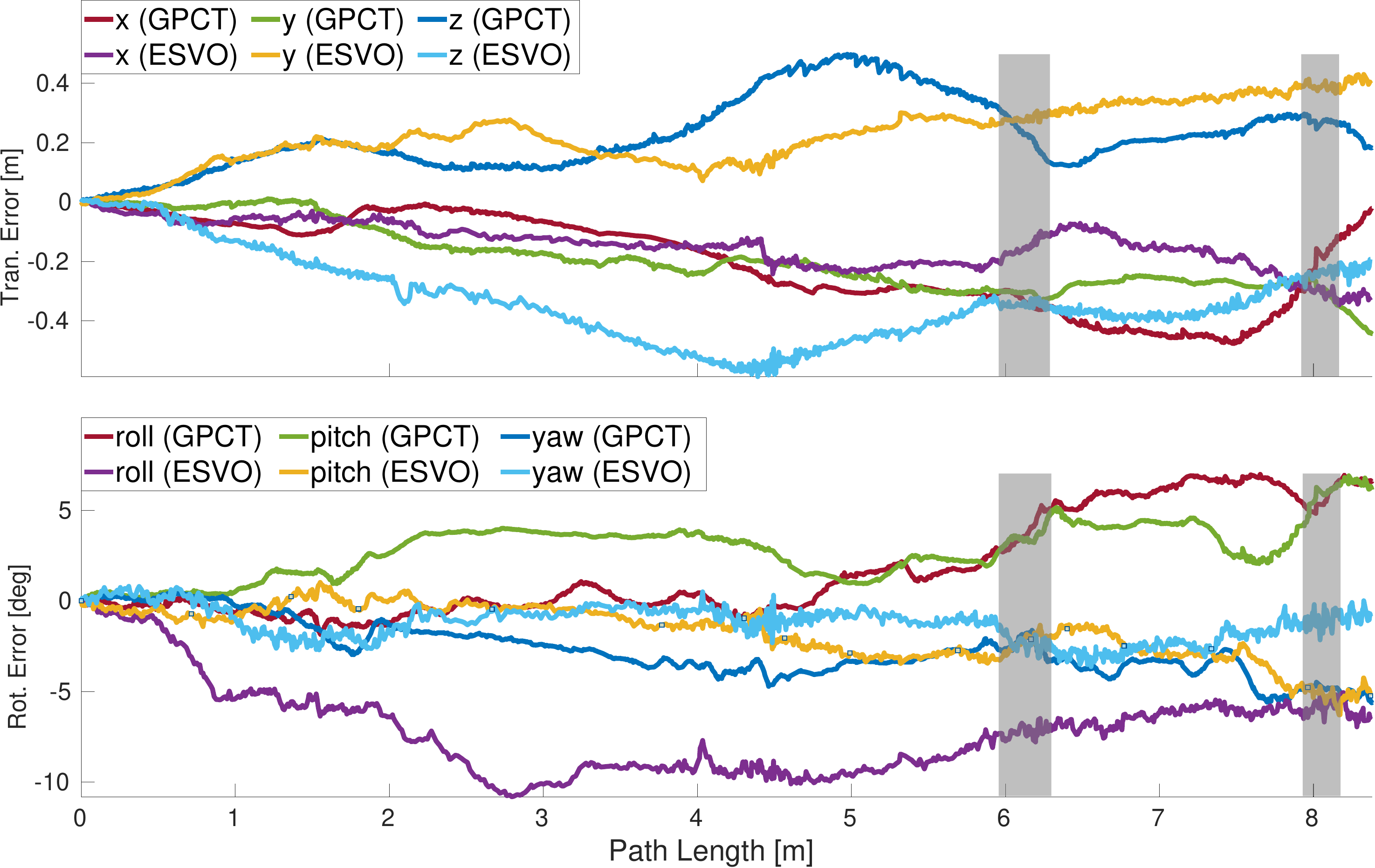}
  \caption{\textit{Indoor1} global error}
  \label{fig:GE_indoor1}
  \end{subfigure}
  \hfill
  \begin{subfigure}{0.49\linewidth}
  \includegraphics[width=1\linewidth]{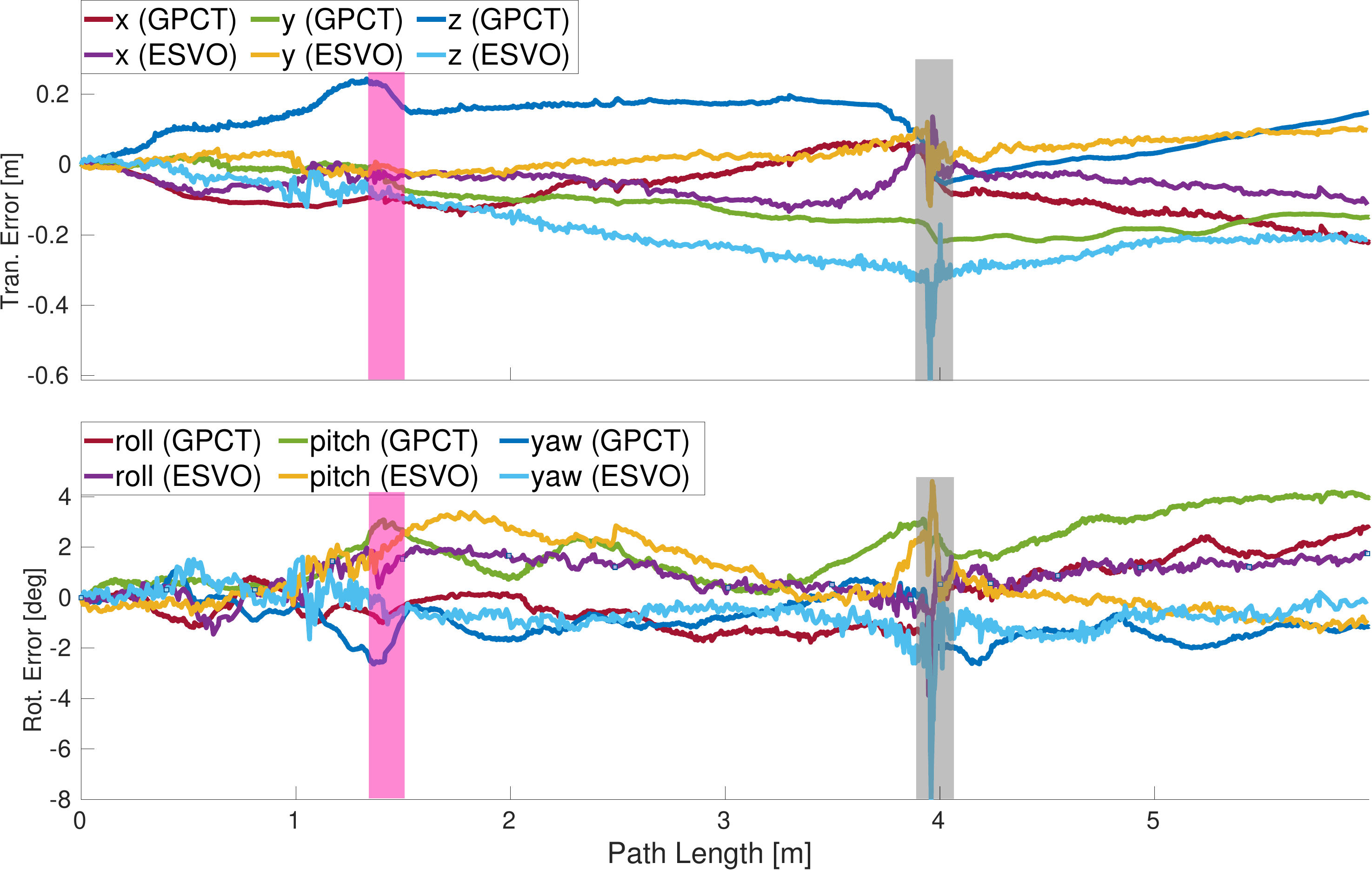}
  \caption{\textit{Indoor3} global error}
  \label{fig:GE_indoor3}
  \end{subfigure}
  \vspace{5pt}
  \\
  \begin{subfigure}{0.49\linewidth}
  \includegraphics[width=1\linewidth]{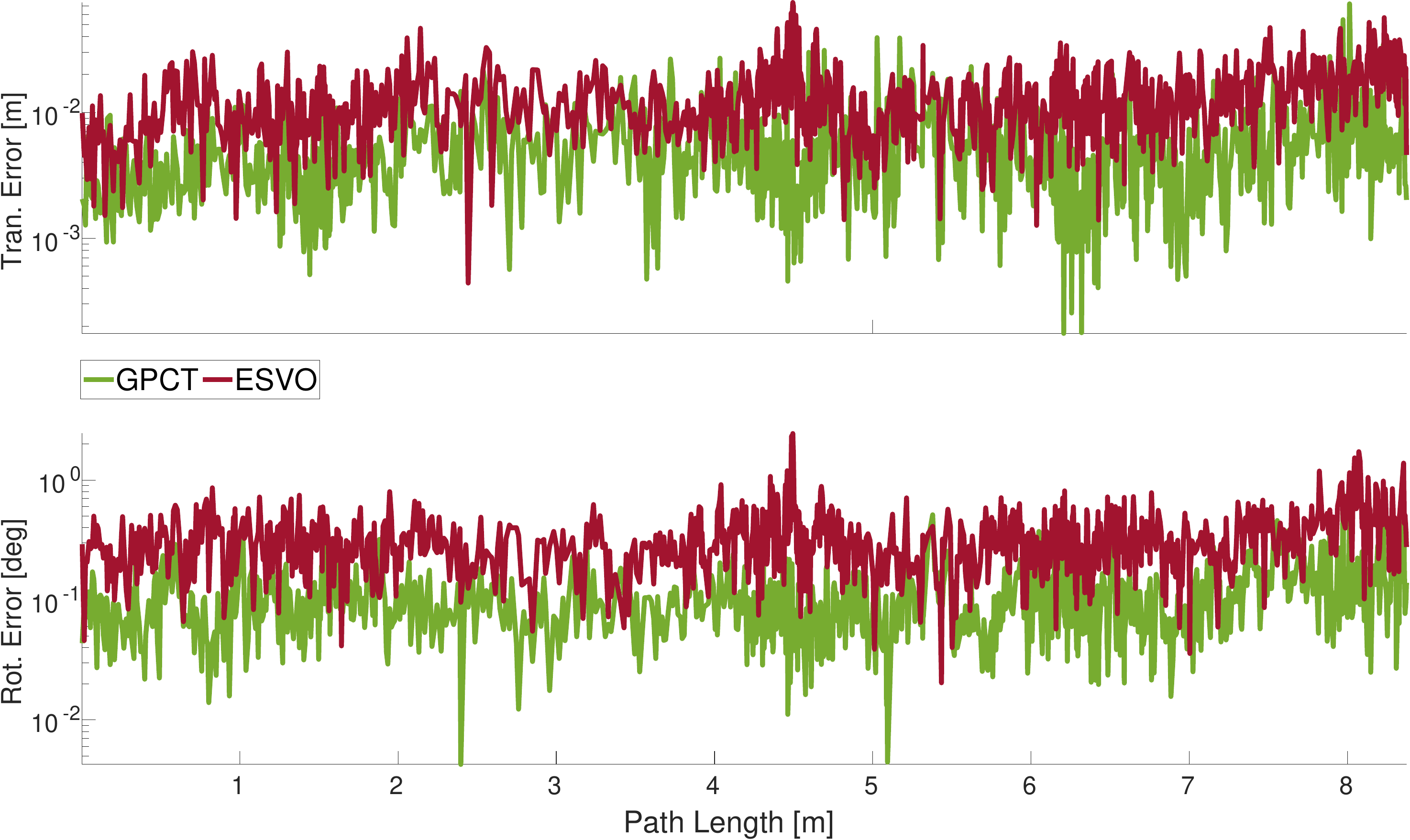}
  \caption{\textit{Indoor1} relative error}
  \label{fig:RE_indoor1}
  \end{subfigure}
   \hspace{1ex}%
  \begin{subfigure}{0.49\linewidth}
  \includegraphics[width=1\linewidth]{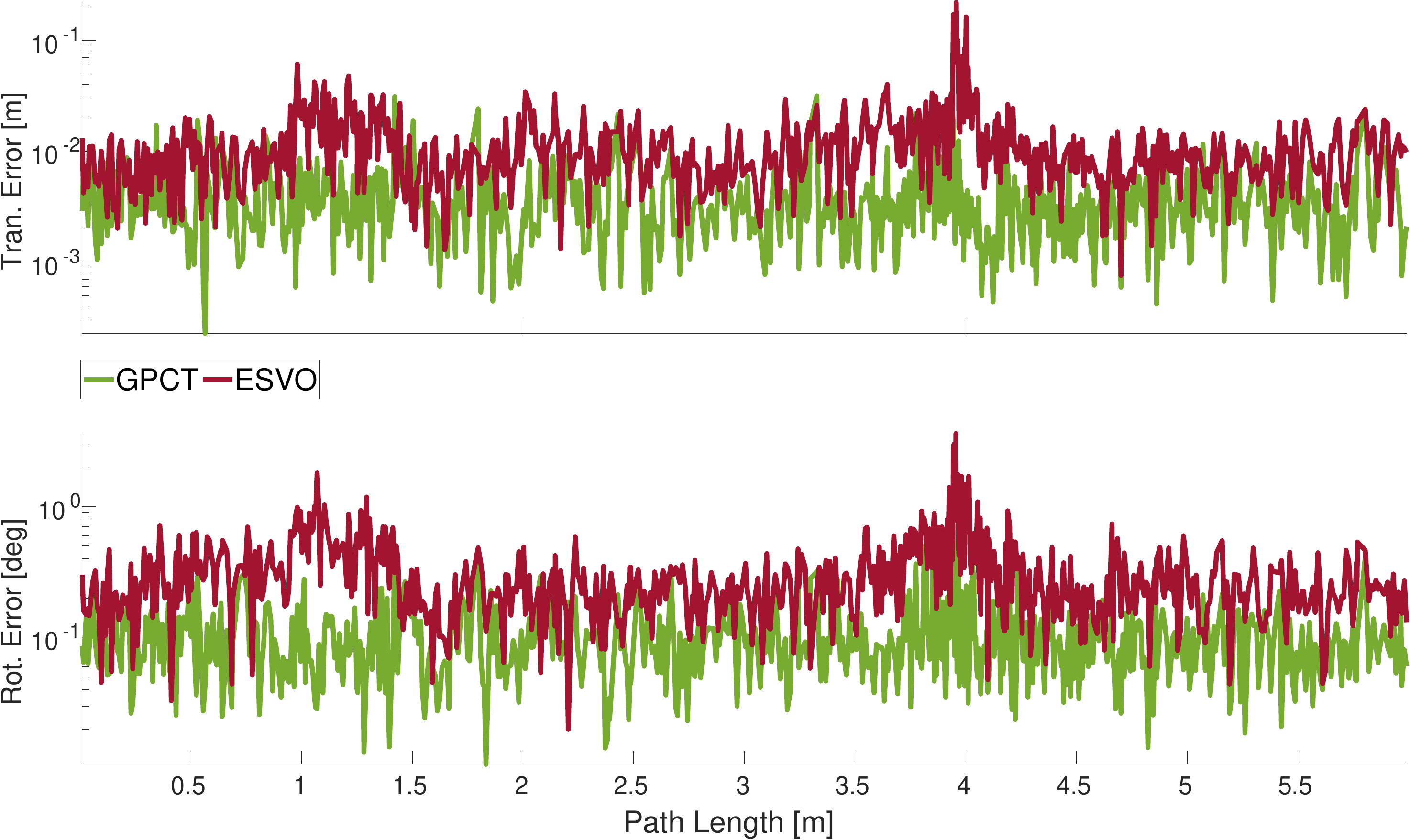}
  \caption{\textit{Indoor3} relative error}
  \label{fig:RE_indoor3}
  \end{subfigure}
  \caption{\footnotesize{The global and relative error of the presented Gaussian process continuous-time approach (GPCT) and ESVO as a function of path length on MVSEC \textit{indoor1} and \textit{indoor3}. The translational and rotational global error are plotted separately and the relative error is plotted as a single $SE(3)$ quantity. Note that the relative error y-axis is a logarithmic scale which suppresses spikes. The grey shaded areas represent complex camera motions that result in poor feature quality. The pink areas indicate featureless regions where the presented technique relies solely on the motion prior.}} \vspace{-1\baselineskip}
  \label{fig:GE_RE}
\end{figure*}

Challenging camera motions are marked in Fig. \ref{fig:GE_indoor3} as pink and grey shaded areas. The grey areas correspond to a back-and-forth camera motion and results in poor performance for both techniques. ESVO loses tracking and has to reinitialize, resulting in large global and relative error spikes (Figs. \ref{fig:GE_indoor3} and \ref{fig:RE_indoor3}). The presented system performs better than ESVO in this area, but feature quality also decreases which increases both global and relative error.

The pink areas correspond to a large rotational motion. Feature tracking  fails for the presented system during this motion but the WNOA prior carries the estimate through without causing significant error.
This demonstrates the robustness of the presented system to feature tracking failure and the potential to improve performance with ongoing research on event-based feature tracking.

\section{CONCLUSIONS}
\label{sec:conclusion}
This paper presents a complete event-based continuous-time VO pipeline that maintains the temporal resolution of event cameras throughout the estimation. This pipeline can use either traditional frame-based or new event-based feature detectors and trackers to generate asynchronous tracklets. These tracklets are filtered for outliers using a motion-compensated RANSAC that accounts for the unique tracklet times. The pipeline estimates a continuous-time trajectory using nonparametric Gaussian process regression with a physically founded WNOA motion prior that can be queried for the camera pose at any time within the estimation window. The system's performance is evaluated on the publicly available MVSEC dataset where it achieves better performance than the publicly available ESVO pipeline, especially in terms of RMS relative error.

\mycomment{
\footnotesize{
\section*{\footnotesize{ACKNOWLEDGMENT}} %
This research was funded by UK Research and Innovation and EPSRC through ACE-OPS: From Autonomy to Cognitive assistance in Emergency OPerationS [EP/S030832/1].
}
}

\begin{spacing}{0.99}
\footnotesize{
\bibliographystyle{ieeetr} %
\bibliography{paper} %
}
\end{spacing}

\end{document}